# A Heuristic Search Algorithm for Solving First-Order MDPs

Eldar Karabaev
Institute for Theoretical Computer Science
Computer Science Department
Technische Universität Dresden
karabaev@tcs.inf.tu-dresden.de

Olga Skvortsova
International Center for Computational Logic
Technische Universität Dresden
skvortsova@iccl.tu-dresden.de

We present a heuristic search algorithm for solving first-order MDPs (FOMDPs). Our approach combines *first-order state abstraction* that avoids evaluating states individually, and *heuristic search* that avoids evaluating all states. Firstly, we apply state abstraction directly on the FOMDP avoiding propositionalization. Such kind of abstraction is referred to as first-order state abstraction. Secondly, guided by an admissible heuristic, the search is restricted only to those states that are reachable from the initial state. We demonstrate the usefullness of the above techniques for solving FOMDPs on a system, referred to as FC-Planner, that entered the probabilistic track of the International Planning Competition (IPC'2004).

## 1 INTRODUCTION

Markov decision processes (MDPs) have been adopted as a representational and computational model for decision-theoretic planning problems in much recent work, e.g., (Barto et al., 1995). The basic solution techniques for MDPs (Boutilier et al., 1999) rely on the dynamic programming (DP) principle. Unfortunately, classical dynamic programming algorithms require explicit enumeration of state space that grows exponentially with the number of variables relevant to the planning domain. Therefore, these algorithms do not scale up to complex AI planning problems.

However, several methods that avoid the explicit state enumeration have been recently developed. One technique, referred to as state abstraction, exploits the structure of the factored MDP representation to solve the problems efficiently, circumventing explicit state space enumeration (Boutilier et al., 1999). Another technique, referred to as heuristic search, restricts the computation to states that are reachable from the initial state (e.g., RTDP (Barto et al., 1995), envelope DP (Dean et al., 1995) and LAO$^*$ (Hansen and Zilberstein, 2001)). One existing approach that combines these both techniques is the symbolic LAO$^*$ algorithm which performs heuristic search symbolically for factored MDPs (Feng and Hansen, 2002). It exploits state abstraction, i.e., manipulates with sets of states instead of individual states. More precisely, following the SPUDD approach (Hoey et al., 1999), all MDP components, value functions, policies, and admissible heuristic functions are compactly represented using algebraic decision diagrams (ADDs). This allows to efficiently perform all computations of the LAO$^*$ algorithm using ADDs.

Following ideas of symbolic LAO$^*$, given an initial state, we use an admissible heuristic to restrict search only to those states that are reachable from the initial state. Moreover, we exploit state abstraction in order to avoid evaluating states individually. Thus, our work is very much in the spirit of symbolic LAO$^*$ but extends it in an important way. Whereas the symbolic LAO$^*$ algorithm starts with propositionalizing, or grounding, the FOMDP and only after that performs state abstraction on its propositionalized version by means of propositional ADDs, we apply state abstraction directly on the structure of the FOMDP avoiding propositionalization. Such kind of abstraction is referred to as *first-order state abstraction*.

Recently, following (Boutilier et al., 2001), we have developed an algorithm, referred to as first-order value iteration algorithm (FOVIA) that exploits first-order state abstraction (Hölldobler and Skvortsova, 2004). The dynamics of an MDP is specified in the Probabilistic Fluent Calculus, that is a first-order language for reasoning about states and actions (Hölldobler and Schneeberger, 1990). More precisely, FOVIA produces a logical representation of value functions and policies by constructing first-order formulae that partition state space into clusters, referred to as *abstract states*. In effect, it performs value iteration on top of these clusters, obviating the need in the explicit state enumeration. This allows problems that are represented in first-order terms to be solved without requiring explicit state enumeration or propositionalization.

Indeed, FOMDP's propositionalization is very impractical: the number of propositions grows considerably with the number of domain objects and relations. This has a dramatic impact on the complexity of the algorithms that depends directly on the number of propositions. Moreover, as soon as the universe of objects is infinite, these algorithms cannot be made to work. Finally, systems for solving FOMDPs that rely on state propositionalization also perform action propositionalization which is problematic in first-order domains, because the number of ground actions also grows dramatically with domain size.

In this paper, we address these difficulties by proposing an approach for solving FOMDPs that combines first-order state abstraction and heuristic search in a novel way exploiting the power of logical representations. Our algorithm can be viewed as a first-order generalization of LAO*, in which our contribution is to show how to perform heuristic search for first-order MDPs, circumventing their propositionalization. In fact, we show how to improve the performance of symbolic LAO* by providing a compact first-order MDP representation using Probabilistic Fluent Calculus instead of propositional ADDs. On the other hand, our approach can be considered as a way to improve the efficiency of the FOVIA algorithm by using heuristic search together with the symbolic dynamic programming.

## 2 FIRST-ORDER REPRESENTATION OF MDPS

Recently, several representations for propositionally-factored MDPs have been proposed, including dynamic Bayesian networks (Boutilier et al., 1999) and ADDs (Hoey et al., 1999). For instance, the SPUDD algorithm (Hoey et al., 1999) has been used to solve MDPs with hundreds of millions of states optimally, producing logical descriptions of value functions that involve only hundreds of distinct values. This work demonstrates that large MDPs, described in a logical fashion, can often be solved optimally by exploiting the logical structure of the problem. Meanwhile, many realistic planning domains are best represented in first-order terms. However, most existing implemented solutions for first-order MDPs rely on propositionalization, i.e., eliminate all variables at the outset of a solution attempt by instantiating terms with all possible combinations of domain objects. This technique is very impractical because the number of propositions grows dramatically with the number of domain objects and relations. For example, the goal statement $\exists X_0 \ldots X_7.\ red(X_0) \wedge green(X_1) \wedge \ldots \wedge blue(X_7) \wedge Tower(X_0, \ldots, X_7)$ in a colored Blocksworld problem of only eight blocks and three colors (where along with the unique identifier each block is assigned a specific color) could result in up to 144 different combinations of blocks, when grounded.

To address these difficulties, we propose a concise representation of FOMDPs within Probabilistic Fluent Calculus that is a logical approach to modelling dynamically changing systems based on first-order logic.

### 2.1 MDPs

A Markov decision process (MDP), is a tuple $(\mathcal{Z}, \mathcal{A}, \mathcal{P}, \mathcal{R}, \mathcal{C})$, where $\mathcal{Z}$ is a finite set of states, $\mathcal{A}$ is a finite set of actions, and $\mathcal{P} : \mathcal{Z} \times \mathcal{Z} \times \mathcal{A} \to [0, 1]$, written $\mathcal{P}(z'|z, a)$, specifies transition probabilities. In particular, $\mathcal{P}(z'|z, a)$ denotes the probability of ending up at state $z'$ given that the agent was in state $z$ and action $a$ was executed. $\mathcal{R} : \mathcal{Z} \to \Re$ is a real-valued reward function associating with each state $z$ its immediate utility $\mathcal{R}(z)$. $\mathcal{C} : \mathcal{A} \to \Re$ is a real-valued cost function associating a cost $\mathcal{C}(a)$ to each action $a$. A sequential decision problem consists of an MDP and is the problem of finding a policy $\pi : \mathcal{Z} \to \mathcal{A}$ that maximizes the total expected discounted reward received when executing the policy $\pi$ over an infinite (or indefinite) horizon. The value of a state $z$ with respect to a policy $\pi$ is defined recursively as:

$$V_\pi(z) = \mathcal{R}(z) + \mathcal{C}(\pi(z)) + \gamma \sum_{z' \in \mathcal{Z}} \mathcal{P}(z'|z, \pi(z)) V_\pi(z'),$$

where $0 \leq \gamma \leq 1$ is a discount factor. We take $\gamma$ equal to 1 for indefinite-horizon problems only, i.e., when a goal is reached the system enters an absorbing state in which no further rewards or costs are accrued. The optimal value function $V^*$ satisfies:

$$V^*(z) = \mathcal{R}(z) + \max_{a \in \mathcal{A}} \{\mathcal{C}(a) + \gamma \sum_{z' \in \mathcal{Z}} \mathcal{P}(z'|z, a) V^*(z')\},$$

for each $z \in \mathcal{Z}$.

### 2.2 PROBABILISTIC FLUENT CALCULUS

Fluent Calculus ($\mathcal{FC}$) was originally set up as a first-order logic program with equality using SLDE-resolution as sole inference rule (Hölldobler and Schneeberger, 1990). The Probabilistic Fluent Calculus ($\mathcal{PFC}$) is an extension of the original $\mathcal{FC}$ for expressing planning domains with probabilistic effects.

**States:** Formally, let $\Sigma$ denote a set of function symbols. We distinguish two function symbols in $\Sigma$, namely $\circ/2$ which is associative, commutative, and admits the unit element, and a constant 1. Let $\Sigma_- = \Sigma \setminus \{\circ, 1\}$. Non-variable $\Sigma_-$-terms are called *fluents*.

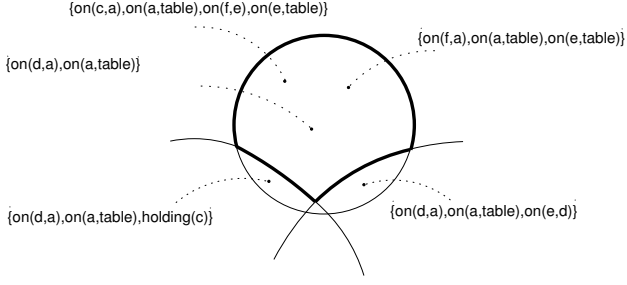

Figure 1: The interpretation of a *CN-state* $Z = (on(X,a) \circ on(a, table), \{holding(X'), on(Y,X)\})$.

For example, $on(X, table)$ is a fluent meaning that some block $X$ is on the table. *Fluent terms* are defined inductively as follows: 1 is a fluent term; each fluent is a fluent term; $F \circ G$ is a fluent term, if $F$ and $G$ are fluent terms. For example, $on(b, table) \circ holding(X)$ is a fluent term denoting that the block $b$ is on the table and some block $X$ is in the robot's gripper. We denote a set of fluents as $\mathcal{F}$ and a set of fluent terms as $\mathcal{L}^{\mathcal{F}}$, resp. A *state* is defined by a pair $(P, \mathcal{N})$, where $P \in \mathcal{L}^{\mathcal{F}}$ and $\mathcal{N} \in 2^{\mathcal{L}^{\mathcal{F}}}$. We refer to states in $\mathcal{PFC}$ as *CN-states*, where $C$ stands for conjunction and $N$ for negation as they are the only connectives that are allowed in state descriptions. We denote *CN-states* by $Z, Z_1, Z_2$ etc. and the set of *CN-states* by $\mathcal{L}_{CN}$.

Let $\cdot^M$ be a mapping from fluent terms to multisets of fluents, which can be formally defined as follows: $1^M = \dot{\{\}}$ or $F^M = \dot{\{F\}}$, if $F \in \mathcal{F}$, or $(F \circ G)^M = F^M \dot{\cup} G^M$, where $F, G \in \mathcal{L}^{\mathcal{F}}$ and $\dot{\cup}$ is a multiset union. The interpretation over $\mathcal{F}$, denoted as $\mathcal{I}$, is the pair $(\Delta, \cdot^{\mathcal{I}})$, where the domain $\Delta$ is a set of all finite multisets of ground fluents from $\mathcal{F}$; and an interpretation function $\cdot^{\mathcal{I}}$ which assigns to each fluent term $F$ a set $F^{\mathcal{I}} \subseteq \Delta$ and to each *CN-state* $Z = (P, \mathcal{N})$ a set $Z^{\mathcal{I}} \subseteq \Delta$ as follows:

$F^{\mathcal{I}} = \{d \in \Delta | \exists \theta.(F\theta)^M \dot{\subseteq} d\}$
$Z^{\mathcal{I}} = \{d \in \Delta | \exists \theta.(P\theta)^M \dot{\subseteq} d \land \forall N \in \mathcal{N}.d \notin (N\theta)^{\mathcal{I}}\},$

where $\dot{\subseteq}$ is a submultiset relation and $\theta$ is a substitution. For example, the interpretation of a *CN-state*

$Z = (on(X,a) \circ on(a, table), \{holding(X'), on(Y,X)\})$

that can be read: There exists a block $X$ that is on the block $a$ which is on the table and there exists no such block $X'$ that the robot holds and there is no such block $Y$ that is on $X$, is depicted on Figure 1.

Since $Z^{\mathcal{I}}$ contains all such finite multisets of ground fluents that satisfy the $P$-part and do not satisfy any of the elements of the $\mathcal{N}$-part, we subtract all multisets that belong to each of $N_i \in \mathcal{N}$ from the set of multisets that correspond to the $P$-part. Thus, the bolded area on Figure 1 contains exactly those multisets that do satisfy the $P$-part of $Z$ and none of the elements of its $\mathcal{N}$-part. For example, a real-world state $\dot{\{}on(d,a), on(a, table)\dot{\}}$ belongs to $Z^{\mathcal{I}}$, whereas $\dot{\{}on(d,a), on(a, table), holding(c)\dot{\}}$ does not. In other words, *CN-states* are characterized by means of conditions that must hold in each ground instance thereof and, thus, they represent clusters of real-world, or individual, states. In this way, *CN-states* embody a form of state space abstraction and, hence, can be treated as abstract states. Such kind of abstraction is referred to as *first-order state abstraction*.

**Actions:** *Actions* are first-order terms leading with an action function symbol. For example, the action of picking up some block $X$ from another block $Y$ might be denoted as $pickup(X,Y)$. Formally, let $N_a$ denote a set of action names disjoint with $\Sigma$. An *action space* is a tuple $\mathcal{A} = (A, Pre, Eff)$, where $A$ is a set of terms of the form $a(p_1, \ldots, p_n)$, referred to as *actions*, with $a \in N_a$ and each $p_i$ being either a variable, or a constant; $Pre : A \to \mathcal{L}_{CN}$ is a *precondition* of $a$; and $Eff : A \to \mathcal{L}_{CN}$ is an *effect* of $a$.

So far, we have described deterministic actions only. But actions in $\mathcal{PFC}$ may have probabilistic effects as well. In order to model these, we decompose a stochastic action into deterministic primitives under nature's control, referred to as *nature's choices*. We use a relation symbol $choice/2$ to model nature's choice. Consider the action $pickup(X,Y)$:

$choice(pickup(X,Y), A) \leftrightarrow$
$\quad (A = pickupS(X,Y) \lor A = pickupF(X,Y)),$

where $pickupS(X,Y)$ and $pickupF(X,Y)$ define two nature's choices for action $pickup(X,Y)$, viz., that it is successfully executed or fails. For example, the nature's choice $pickupS$ can be defined as follows:

$Pre(pickupS(X,Y)) := (on(X,Y) \circ e, \{on(W,X)\})$
$Eff(pickupS(X,Y)) := (holding(X), \{on(X,Y)\}),$

where the fluent $e$ denotes the empty robot's gripper. For simplicity, we denote the set of nature's choices of an action $a$ as $Ch(a) := \{a_j | choice(a, a_j)\}$. Please note that nowhere do these action descriptions restrict the domain of discourse to some prespecified set of blocks. Moreover, domains with infinitely many individuals can be effortlessly represented in this way as well. For each of nature's choices $a_j$ associated with an action $a$ we define the probability $prob(a_j, a, Z)$ denoting the probability with which one of nature's choices $a_j$ is chosen in a state $Z$. For example,

$prob(pickupS(X,Y), pickup(X,Y), Z) = .75$

states that the probability for the successful execution of the $pickup$ action in state $Z$ is .75.

In the next step, we define the reward function for each state. We give a reward of 500 to all states in which some block $X$ is on block $a$ and 0, otherwise:

$$reward(Z) = 500 \leftrightarrow Z \sqsubseteq (on(X,a), \emptyset)$$
$$reward(Z) = 0 \leftrightarrow Z \not\sqsubseteq (on(X,a), \emptyset) ,$$

where $\sqsubseteq$ denotes the subsumption relation: A CN-state $Z_1$ subsumes a CN-state $Z_2$, written $Z_2 \sqsubseteq Z_1$, iff $Z_2^\mathcal{I} \subseteq Z_1^\mathcal{I}$. One should observe that we have specified the reward function without explicit state enumeration. Instead, the state space is divided into two abstract states depending on whether or not, a block $X$ is on block $a$. Likewise, value functions can be specified with respect to the abstract states only. This is in contrast to classical DP algorithms, in which the states are explicitly enumerated. Action costs can be analogously defined as follows:

$$cost(pickup(X,Y)) = -3$$

penalizing the execution of the *pickup*-action with the value of 3.

**Inference Mechanism:** In this section, we show how to perform regression and progression directly on abstract states avoiding propositionalization. Let $Z = (P, \mathcal{N}) \in \mathcal{L}_{CN}$, $a(p_1, \ldots, p_n)$ be an action with parameters $p_1, \ldots, p_n$, preconditions $Pre(a) = (P_p, \mathcal{N}_p)$ and effects $Eff(a) = (P_e, \mathcal{N}_e)$. Let $\theta$ be a substitution. An action $a(p_1, \ldots, p_n)$ is *forward applicable*, or simply *applicable*, to $Z$ with $\theta$, denoted as $forward(Z, a, \theta)$, if the following condition holds:

$(P_p\theta)^M \dot{\subseteq} P^M \wedge$
$\forall N_p \in \mathcal{N}_p. \exists N \in \mathcal{N}. \exists \sigma. ((P \circ N)\sigma)^M \dot{\subseteq} ((P \circ N_p)\theta)^M .$

In other words, the above statement guarantees that $Z$ contains both positive and negative preconditions of the action $a$. Similarly, an action $a(p_1, \ldots, p_n)$ is *backward applicable* to $Z$ with $\theta$, denoted as $backward(Z, a, \theta)$, if $Z$ contains both positive and negative effects of $a$, i.e.:

$(P_e\theta)^M \dot{\subseteq} P^M \wedge$
$\forall N_e \in \mathcal{N}_e. \exists N \in \mathcal{N}. \exists \sigma. ((P \circ N)\sigma)^M \dot{\subseteq} ((P \circ N_e)\theta)^M .$

If an action $a$ is forward applicable to $Z$ with $\theta$ then $Z' = ((P')^{-M}, \mathcal{N}')$, where

$$P' := P^M \dot{\setminus} (P_p\theta)^M \dot{\cup} (P_e\theta)^M \qquad (1)$$
$$\mathcal{N}' := \mathcal{N}\theta \setminus \mathcal{N}_p\theta \cup \mathcal{N}_e\theta$$

is referred to as the *a*-**successor** of $Z$ with $\theta$ and denoted as $succ(Z, a, \theta)$. Similarly, if an action $a$ is backward applicable to $Z$ with $\theta$ then $Z'' = ((P'')^{-M}, \mathcal{N}'')$, where

$$P'' := P^M \dot{\setminus} (P_e\theta)^M \dot{\cup} (P_p\theta)^M \qquad (2)$$
$$\mathcal{N}'' := \mathcal{N}\theta \setminus \mathcal{N}_e\theta \cup \mathcal{N}_p\theta$$

is referred to as the *a*-**predecessor** of $Z$ with $\theta$ and denoted as $pred(Z, a, \theta)$. For example, consider the action $pickupS(X, Y)$ as defined above, take $Z = (P, \mathcal{N}) = (on(b, table) \circ on(X_1, b) \circ e, \{on(X_2, X_1)\})$. The action $pickupS(X, Y)$ is forward applicable to $Z$ with $\theta = \{X \mapsto X_1, Y \mapsto b\}$. Thus, $Z' = succ(Z, pickupS(X, Y), \theta) = ((P')^{-M}, \mathcal{N}')$ with

$$P' = \{on(b, table), holding(X_1)\} \quad \mathcal{N}' = \{on(X_1, b)\} .$$

In effect, Equations 1 and 2 comprise the inference mechanism for computing predecessor and successor abstract states. This mechanism operates on abstract states directly, instead of evaluating individual states.

## 3 FIRST-ORDER LAO*

We present a generalization of symbolic LAO* algorithm (Feng and Hansen, 2002), referred to as first-order LAO* (FOLAO*), for solving FOMDPs. Symbolic LAO* is a heuristic search algorithm that exploits state abstraction for solving factored MDPs. Given an initial state, symbolic LAO* uses an admissible heuristic to focus computation on the parts of the state space that are reachable from the initial state. Moreover, it specifies MDP components, value functions, policies, and admissible heuristics using propositional ADDs. This allows symbolic LAO* to manipulate sets of states instead of individual states.

Despite the fact that symbolic LAO* shows an advantageous behaviour in comparison to non-symbolic LAO* that evaluates states individually, it suffers from an important drawback. While solving FOMDPs symbolic LAO* performs problem propositionalization. This approach is impractical for large FOMDPs and is hardly made to work when the domain becomes infinite. Our intention is to show how to improve the performance of symbolic LAO* by providing a compact first-order representation of MDPs so that the heuristic search can be performed without propositionalization. More precisely, we propose to switch the representational formalism for FOMDPs in symbolic LAO* from propositional ADDs to Probabilistic Fluent Calculus. The FOLAO* algorithm is presented on Figure 2.

As symbolic LAO*, FOLAO* has two phases that alternate until a complete solution is found, which is guaranteed to be optimal. First, it expands the best partial policy and evaluates the states on its fringe using an admissible heuristic function. Then it performs dynamic programming on the states visited by the best partial policy, to update their values and possibly revise the current best partial policy.

In the policy expansion step, we perform reachability analysis to find the set of states $F$ that have not yet

```
policyExpansion(π, S⁰, G)
  E := F := ∅
  from := S⁰
  repeat
    to :=   ⋃       ⋃      {succ(Z, aⱼ, θ)},
          Z∈from  aⱼ∈Ch(a)
    where (a, θ) := π(Z)
    F := F ∪ (to − G)
    E := E ∪ from
    from := to ∩ G − E
  until (from = ∅)
  E := E ∪ F
  G := G ∪ F
  return (E, F, G)

FOVIA(𝒜, prob, reward, cost, γ, E, V)
  repeat
    V' := V
    loop for each Z ∈ E
      loop for each a ∈ 𝒜
        loop for each θ such that forward(Z, a, θ)
          Q(Z, a, θ) := reward(Z) + cost(a)+
            γ  ∑     prob(aⱼ, a, Z) · V'(succ(Z, aⱼ, θ))
              aⱼ∈Ch(a)
        end loop
      end loop
      V(Z) := max   Q(Z, a, θ)
              (a,θ)
    end loop
    r := ‖V − V'‖
  until stopping criterion
  π := extractPolicy(V)
  return (V, π, r)

FOLAO*(𝒜, prob, reward, cost, γ, S⁰, h, ε)
  V := h
  G := ∅
  For each Z ∈ S⁰, initialize π with an arbitrary action
  repeat
    (E, F, G) := policyExpansion(π, S⁰, G)
    (V, π, r) := FOVIA(𝒜, prob, reward, cost, γ, E, V)
  until (F = ∅) and r ≤ ε
  return (π, V)
```

Figure 2: First-order LAO* algorithm.

been expanded, but are reachable from the set of initial states $S^0$ by following the partial policy $\pi$. The set of states $G$ contains states that have been expanded so far. By expanding a partial policy we mean that it will be defined for a larger set of states in the dynamic programming step. In symbolic LAO*, reachability analysis on ADDs is performed by means of the *image* operator, taken from the area of symbolic model checking, that computes the set of successor states *to* following the best current policy. Whereas, in FOLAO*, we apply the *succ*-operator, defined in Equation 1. One should observe that since the reachability analysis in FOLAO* is performed on *CN-states* that are defined as first-order entities, the reasoning about successor states is kept on the first-order level.

Whereas the symbolic LAO* should first instantiate $S^0$ with all possible combinations of objects, in order to be able to perform computations using propositional ADDs later on.

In contrast to symbolic LAO*, where the dynamic programming step is performed using a modified version of SPUDD, we employ a modified first-order value iteration algorithm (FOVIA) (Hölldobler and Skvortsova, 2004). The original FOVIA performs value iteration over the entire state space. We modify it so that it computes on states that are reachable from the initial states, more precisely, on the set of states $E$ that are visited by the best current partial policy. In this way, we improve the efficiency of the original FOVIA by using the reachability analysis together with the symbolic dynamic programming. FOVIA produces a $\mathcal{PFC}$ representation of value functions and policies by constructing first-order formulae that partition state space into abstract states specified as *CN-states*. In effect, it performs value iteration on top of *CN-states*, obviating the need in the explicit state enumeration.

Given a FOMDP and a value function represented in $\mathcal{PFC}$, FOVIA returns the best partial value function $V$, the best partial policy $\pi$ and the residual $r$. In order to update the values of the states $Z$ in $E$, we assign the values from the current value function to the successors of $Z$. We compute successors with respect to all nature's choices $a_j$. The residual $r$ is computed as a largest absolute value of the difference between the current and the newly computed value functions $V'$ and $V$, resp. Extraction of a best partial policy $\pi$ is straightforward: one simply needs to extract the maximizing actions from the best partial value function $V$. The elegance of the first-order representation of MDPs allows us to effortlessly restrict the set of states to reachable ones, without the need to perform anything similar to masking of ADDs, as in symbolic LAO*.

As symbolic LAO*, FOLAO* converges to an $\varepsilon$-optimal policy when three conditions are met: its current policy does not have any unexpanded states, the residual $r$ is less than the predefined threshold $\varepsilon$, and the value function is initialized with an admissible heuristic. The convergence proofs for the symbolic LAO* carry over in a straightforward way to FOLAO* (Hansen and Zilberstein, 2001).

At the beginning of FOLAO*, we initialize the value function with an admissible heuristic function $h$ that focuses the search on a subset of reachable states. A simple way to create admissible heuristic is to use dynamic programming to create an approximate value function. Therefore, in order to create an admissible heuristic $h$ in FOLAO*, we perform several iterations of the original FOVIA. We started the algorithm on an initial value function that is admissible. Since each step of FOVIA preserves admissibility, the resulting value function is admissible as well. The initial value function assigns the goal reward to each state thereby overestimating the optimal value since the goal reward is the maximal possible reward.

Since all computations of FOLAO* are performed on *CN-states* instead of individual states, FOMDPs are solved avoiding explicit state and action enumeration and propositionalization. The reasoning on first-order level leads to better performance of FOLAO* in comparison to symbolic LAO*, as shown in the next section.

## 4 EXPERIMENTAL EVALUATION

We demonstrate the advantages of combining the heuristic search together with first-order state abstraction on a system, referred to as FCPlanner, that has successfully entered the probabilistic track of the competition IPC'2004. The experimental results were all obtained using Linux RedHat machine running at 3.4GHz Pentium IV with 3Gb of RAM.

In Table 1, we present the performance comparison of FCPlanner (denoted as FCP) together with symbolic LAO* (denoted as LAO*) on examples taken from the colored Blocksworld (BW) scenario that was introduced during IPC'2004. The results and the input problems can be found at http://www.wv.inf.tu-dresden.de/~olga/comparison/. Colored BW problems were of our main interest since they were the only ones represented in first-order terms and hence the only ones that allowed us to make use of the first-order state abstraction. These problems differ from the classical BW ones in that, along with the unique identifier, each block is assigned a specific color. A goal formula, specified in first-order terms, provides an arrangement of colors instead of an arrangement of blocks.

At the outset of solving a colored BW problem, symbolic LAO* starts with grounding its components, namely, the goal statement and actions. Only after that, the abstraction using propositional ADDs is applied. Whereas, FCPlanner performs first-order abstraction on a colored BW problem directly, avoiding unnecessary grounding. In the following, we show how an abstraction technique affects the computation of a heuristic function. To create an admissible heuristic, FCPlanner and symbolic LAO* perform twenty iterations of FOVIA and an approximate value iteration algorithm similar to APRICODD (St-Aubin et al., 2000), resp. The columns labelled H.time and NAS show the time needed for computing a heuristic function and the number of abstract states it covers, resp. In comparison to FCPlanner, symbolic LAO* needs to evaluate less abstract states in the heuristic function but takes considerably more time. One can conclude that abstract states in symbolic LAO* enjoy more complex structure than those in FCPlanner.

In order to compare the heuristic accuracy, we present in column labelled NGS the number of ground states that the heuristic assigns non-zero values to. One can see that the heuristics returned by FCPlanner and symbolic LAO* are of close accuracy. But FCPlanner takes much less time to compute it. This reflects the advantage of the plain first-order abstraction in comparison to the marriage of propositionalization with abstraction using propositional ADDs. In some examples, we gain several orders of magnitude in H.time.

The column labelled Total time presents the time needed to solve a problem. During this time, a planner must execute 30 runs from an initial state to a goal state. A one-hour block is allocated for each problem. We note that, in comparison to FCPlanner, the time required by heuristic search in symbolic LAO* (i.e., difference between Total time and H.time) grows considerably faster in the size of the problem. This reflects the potential in employing first-order abstraction instead of abstraction based on propositional ADDs during heuristic search.

The average reward obtained over 30 runs, shown in column Total av. reward, is the planner's evaluation score. The reward value close to 500 simply indicates that a planner found a reasonably good policy. As the number of blocks B increases by 1, the running time for symbolic LAO* increases in 10 times. Thus, it could not scale to problems of the size greater than seven blocks. This is in contrast to FCPlanner that could solve problems of seventeen blocks. We could not analyze the behaviour of FCPlanner on larger problems because these could not be loaded into the current evaluation software that relies on propositionalization. We note that the number of colors C in a problem affects the efficiency of an abstraction technique. In FCPlanner, as C decreases, the abstraction rate increases which in turn is reflected by the dramatic decrease of runtime. The opposite holds for symbolic LAO*.

In addition, we compare FCPlanner with its two variants. The first one, denoted as FOVIA, performs no heuristic search at all. But rather, it employs FOVIA to compute the $\varepsilon$-optimal total value function from which a policy is extracted. The second one, denoted as FCP⁻, performs 'trivial' heuristic search starting with an initial value function as an admissible heuristic. As expected, FCPlanner that combines heuristic search and FOVIA has demonstrated an advantage over plain FOVIA and trivial heuristic search. These results illustrate the significance of heuristic search in general (FCP vs. FOVIA) and importance of heuristic accuracy, in particular (FCP vs. FCP⁻). Even more, FOVIA and FCP⁻ do not scale to problems of the size greater than seven blocks.

Table 1: Performance comparison of FCPlanner (denoted as FCP) and symbolic LAO* (denoted as LAO*)

| Problem | | Total av. reward, ≤500 | | | | Total time, sec. | | | | H.time, sec. | | NAS | | NGS, ×10³ | |
|---|---|---|---|---|---|---|---|---|---|---|---|---|---|---|---|
| B | C | LAO* | FCP | FOVIA | FCP⁻ | LAO* | FCP | FOVIA | FCP⁻ | LAO* | FCP | LAO* | FCP | LAO* | FCP |
| 5 | 4 | 494 | 494 | 494 | 494 | 22.3 | 22.0 | 23.4 | 31.1 | 8.7 | 4.2 | 35 | 410 | 0.86 | 0.82 |
|   | 3 | 496 | 495 | 495 | 496 | 23.1 | 17.8 | 22.7 | 25.1 | 9.5 | 1.3 | 34 | 172 | 0.86 | 0.68 |
|   | 2 | 496 | 495 | 495 | 495 | 27.3 | 11.7 | 15.7 | 16.5 | 12.7 | 0.3 | 32 | 55 | 0.86 | 0.66 |
| 6 | 4 | 493 | 493 | 493 | 493 | 137.6 | 78.5 | 261.6 | 285.4 | 76.7 | 21.0 | 68 | 1061 | 7.05 | 4.24 |
|   | 3 | 493 | 492 | 493 | 492 | 150.5 | 33.0 | 119.1 | 128.5 | 85.0 | 9.3 | 82 | 539 | 7.05 | 6.50 |
|   | 2 | 495 | 494 | 495 | 496 | 221.3 | 16.6 | 56.4 | 63.3 | 135.0 | 1.2 | 46 | 130 | 7.05 | 6.24 |
| 7 | 4 | 492 | 491 | 491 | 491 | 1644 | 198.1 | 2776 | n/a | 757.0 | 171.3 | 143 | 2953 | 65.9 | 23.6 |
|   | 3 | 494 | 494 | 494 | 494 | 1265 | 161.6 | 1809 | 2813 | 718.3 | 143.6 | 112 | 2133 | 65.9 | 51.2 |
|   | 2 | 494 | 494 | 494 | 494 | 2210 | 27.3 | 317.7 | 443.6 | 1241 | 12.3 | 101 | 425 | 65.9 | 61.2 |
| 8 | 4 | n/a | 490 | n/a | n/a | n/a | 1212 | n/a | n/a | n/a | 804.1 | n/a | 8328 | n/a | 66.6 |
|   | 3 | n/a | 490 | n/a | n/a | n/a | 598.5 | n/a | n/a | n/a | 301.2 | n/a | 3956 | n/a | 379.7 |
|   | 2 | n/a | 492 | n/a | n/a | n/a | 215.3 | 1908 | n/a | n/a | 153.2 | n/a | 2019 | n/a | 1121 |
| 15 | 3 | n/a | 486 | n/a | n/a | n/a | 1809 | n/a | n/a | n/a | 1733 | n/a | 7276 | n/a | $1.2 \cdot 10^7$ |
| 17 | 4 | n/a | 481 | n/a | n/a | n/a | 3548 | n/a | n/a | n/a | 1751 | n/a | 15225 | n/a | $2.5 \cdot 10^7$ |

FCPlanner did not perform well on classical BW problems because these problems were propositional ones and FCPlanner does not yet incorporate optimization techniques applied in modern propositional planners. Table 2 concludes with competition results from IPC'2004 where FCPlanner has shown an advantage over other planners on colored BW problems. The contestants are indicated by their origin. E.g., Dresden - FCPlanner, UMass - symbolic LAO*. The gain of five points in total reward means in average ten actions shorter plan.

## 5 RELATED WORK

We follow the symbolic DP (SDP) approach within Situation Calculus (SC) (Boutilier et al., 2001) in using first-order state abstraction for FOMDPs. One difference is in the representation language: we use $\mathcal{PFC}$ instead of SC. In course of the symbolic value iteration, a state space may contain redundant abstract states that dramatically affect the algorithm's efficiency. In order to achieve computational savings, normalization must be performed to remove these redundancies. However, it was done by hand so far. To the best of our knowledge, the preliminary implementation of the SDP approach within SC uses human-provided rewrite rules for logical simplification. Whereas in (Hölldobler and Skvortsova, 2004), we have developed an automated normalization procedure for FOVIA that is incorporated in the competition version of FCPlanner and brings the computational gain of several orders of magnitude. Another crucial difference is that our algorithm uses heuristic search to limit the number of states for which a policy is computed.

ReBel (Kersting et al., 2004) algorithm relates to FOLAO* in that it also uses a simpler logical language than situation calculus which makes the state space simplification computationally feasible.

In motivation, our approach is closely related to Relational Envelope-based Planning (REBP) that represents MDPs dynamics by a compact set of relational rules and extends the envelope method (Dean et al., 1995) to use these structured dynamics (Gardiol and Kaelbling, 2004). However, REBP performs action groundization first and only after that employs abstraction using equivalence-class sampling. Whereas, FOLAO* directly applies state and action abstraction on the first-order structure of an MDP. In that, REBP is closer to symbolic LAO* than to FOLAO*. Moreover, in contrast to $\mathcal{PFC}$, action descriptions in REBP do not allow negation to appear neither in preconditions nor in effects. In organization, FOLAO*, as symbolic LAO*, is similar to real-time DP (Barto et al., 1995) that is an online search algorithm for MDPs, in contrast to FOLAO*, that works offline. There are several recent inductive approaches to solving FOMDPs (Gretton and Thiebaux, 2004; Fern et al., 2003).

Table 2: Official competition results (total average reward) from IPC'2004 (May, 2004)

| Problem | Canberra | **Dresden** | UMass | Michigan | Purdue$_1$ | Purdue$_2$ | Purdue$_3$ | Caracas | Toulouse |
|---|---|---|---|---|---|---|---|---|---|
| 5 blocks | 494.6 | 496.4 | 0 | 0 | 496.5 | 496.5 | 495.8 | 0 | 0 |
| 8 blocks | 486.5 | 492.8 | 0 | 0 | 486.6 | 486.4 | 487.2 | 0 | 0 |
| 11 blocks | 479.7 | 486.3 | 0 | 0 | 481.3 | 481.5 | 481.9 | 0 | 0 |

# 6 CONCLUSIONS

We have proposed an approach that combines heuristic search and first-order state abstraction for solving FOMDPs more efficiently. Our work can be seen as two-fold: First, we use dynamic programming to compute an approximate value function that serves as an admissible heuristic. Then heuristic search is performed to find an exact solution for those states that are reachable from the initial state. In both phases, we exploit the power of first-order state abstraction in order to avoid evaluating states individually. As results show, our approach breaks new ground in exploring the efficiency of first-order representations in solving MDPs. In comparison to existing propositionalization-based MDP planners, e.g., symbolic LAO*, our solution scales better on larger FOMDPs.

However, there is plenty remaining to be done. We are interested in the question to what extent the optimization techniques applied in modern propositional planners can be combined with first-order state abstraction. In future competitions, we would like to face problems where the goal and/or initial states are only partially defined and where the underlying domain contains infinitely many objects.

The current version of FOLAO* is targeted at the problems that allow for efficient first-order state abstraction. More precisely, these are the problems that can be polynomially translated into $\mathcal{PFC}$. E.g., in the competition colored BW domain existentially-closed goal descriptions were linearly translated into the equivalent $\mathcal{PFC}$ representation. Whereas universally-closed goal descriptions would require full groundization. Thus, the current version of $\mathcal{PFC}$ is less first-order expressive than, e.g., Situation Calculus. In the future, it would be promising to study the extensions of the $\mathcal{PFC}$ language, in particular, to find the trade-off between the $\mathcal{PFC}$'s expressive power and the tractability of solution methods for FOMDPs based on $\mathcal{PFC}$.

**Acknowledgements**

We thank anonymous reviewers for useful comments. Many thanks to Zhengzhu Feng for fruitful discussions. Olga Skvortsova was supported by the grant from the German Research Foundation.